\documentclass[sigconf]{acmart}

\usepackage{amsfonts}
\usepackage{algorithmic}
\usepackage{textcomp}
\usepackage{multirow}
\usepackage{colortbl}
\usepackage{makecell}
\definecolor{lightgraytext}{gray}{0.6}
\definecolor{tablegray}{gray}{0.95}
\AtBeginDocument{%
  }

\copyrightyear{2026}
\acmYear{2026}
\setcopyright{cc}
\setcctype{by}
\acmConference[MM '26]{Proceedings of the 34th ACM International Conference on Multimedia}{November 10--14, 2026}{Rio de Janeiro, Brazil}
\acmBooktitle{Proceedings of the 34th ACM International Conference on Multimedia (MM '26), November 10--14, 2026, Rio de Janeiro, Brazil}
\acmDOI{10.1145/3767308.3835856}
\acmISBN{979-8-4007-2213-4/2026/11}

\begin{document}

\title{VisCo: Leveraging Large Language Models as
Intrinsic Encoders for Visual Token Compression}

\author{Yupeng Zheng}
\affiliation{%
    \department{School of Cyber Science and Technology}
  \department{Anhui Province Key Laboratory of Digital Security}
  \institution{University of Science and Technology of China}
  \city{Hefei}
  \state{Anhui}
  \country{China}
}
\email{yupengzheng@mail.ustc.edu.cn}
\orcid{0009-0009-1057-4132}

\author{Kai Zou}
\affiliation{%
    \department{School of Cyber Science and Technology}
  \department{Anhui Province Key Laboratory of Digital Security}
  \institution{University of Science and Technology of China}
  \city{Hefei}
  \state{Anhui}
  \country{China}
}
\email{kzou@mail.ustc.edu.cn}
\orcid{0009-0006-4376-3667}

\author{Bin Liu}
\correspondingauthor
\affiliation{%
  \department{School of Cyber Science and Technology}
  \department{Anhui Province Key Laboratory of Digital Security}
  \institution{University of Science and Technology of China}
  \city{Hefei}
  \state{Anhui}
  \country{China}
}
\email{flowice@ustc.edu.cn}
\orcid{0000-0002-3977-8800}

\author{Nenghai Yu}
\affiliation{%
  \department{School of Cyber Science and Technology}
  \department{Anhui Province Key Laboratory of Digital Security}
  \institution{University of Science and Technology of China}
  \city{Hefei}
  \state{Anhui}
  \country{China}
}
\email{ynh@ustc.edu.cn}
\orcid{0000-0003-4417-9316}

\renewcommand{\shortauthors}{Yupeng Zheng, Kai Zou, Bin Liu, and Nenghai Yu}

\begin{abstract}
Vision-language models (VLMs) process large numbers of visual tokens, resulting in substantial inference latency and memory overhead. This has motivated extensive research on visual token compression. While training-free strategies rely on heuristic metrics and suffer significant performance degradation under high compression ratios, many training-based methods introduce external compression modules that force the VLM backbone to adapt, incurring substantial retraining cost and compromising VLMs' priors. Effective visual token compression hinges on strong information encoding, a capability already present in pretrained VLMs but underutilized by existing approaches. Motivated by this, we propose VisCo, a training-efficient self-compression framework that reuses the pretrained VLM itself as an intrinsic compressor. VisCo is a parameter-sharing autoencoder that compresses visual information using a small set of memory tokens and transfers hierarchical information from encoding to decoding. Experiments show that VisCo surpasses prior methods across all evaluated compression ratios, with larger gains under more aggressive compression, and remains stable even in the extreme single-token setting. Moreover, when combined with the original visual tokens, the learned memory tokens can even improve the base model, suggesting that VisCo captures complementary representations beyond compression. Code is available at: \href{https://github.com/Zyvpeng/VisCo}{\textcolor{blue}{https://github.com/Zyvpeng/VisCo}}
\end{abstract}


\maketitle

\section{Introduction}
\begin{figure}[htbp]
\centering
\includegraphics[width=\columnwidth]{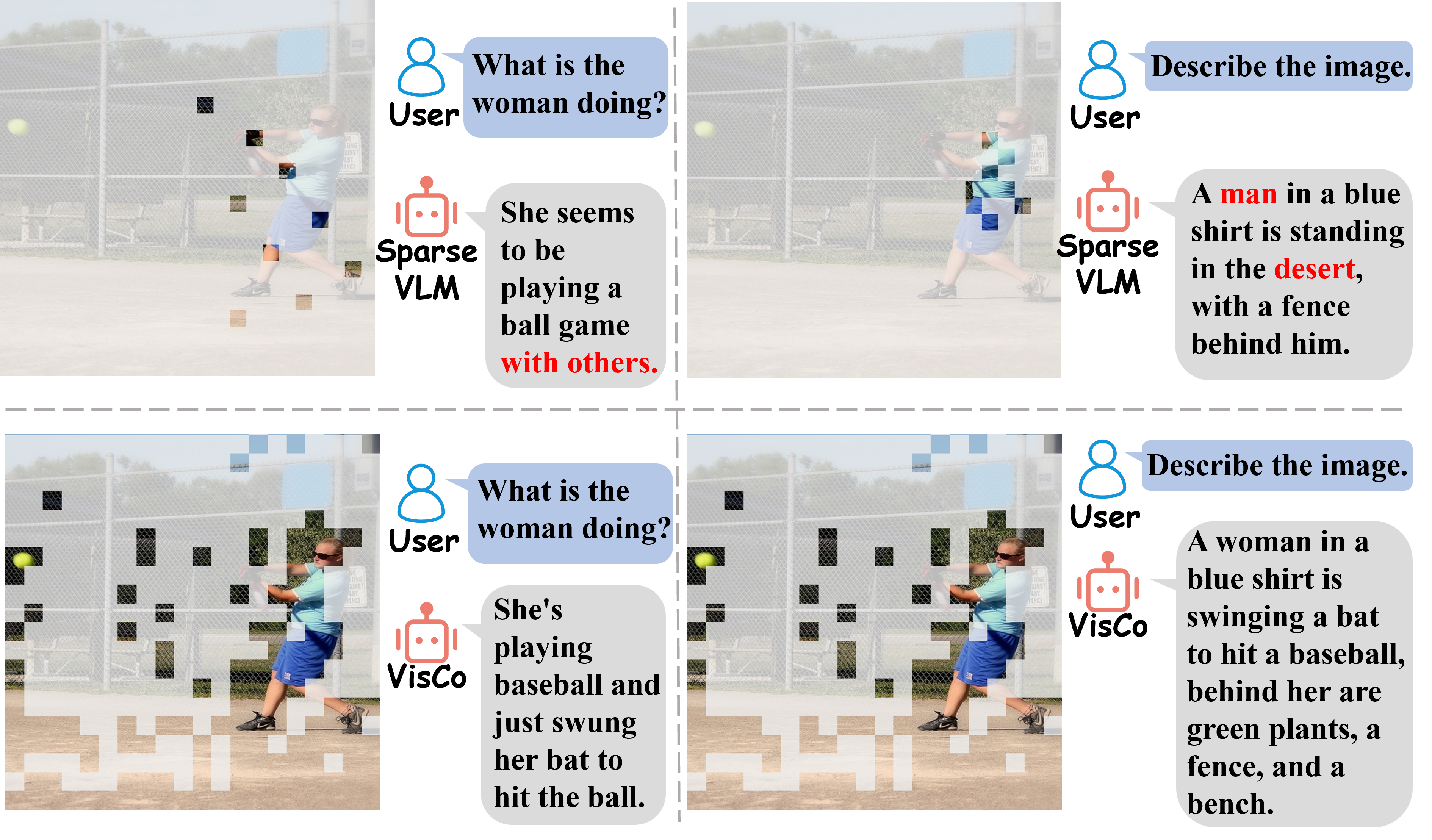}
\caption{Comparison of SparseVLM and VisCo on Qwen2-VL-7B with 9 retained tokens. Top: tokens retained by SparseVLM for two questions. Bottom: the 100 visual tokens receiving the highest attention from VisCo’s memory tokens.}
\label{fig0}
\end{figure}

\begin{figure*}[htbp]
    \centering
    \includegraphics[width=\textwidth]{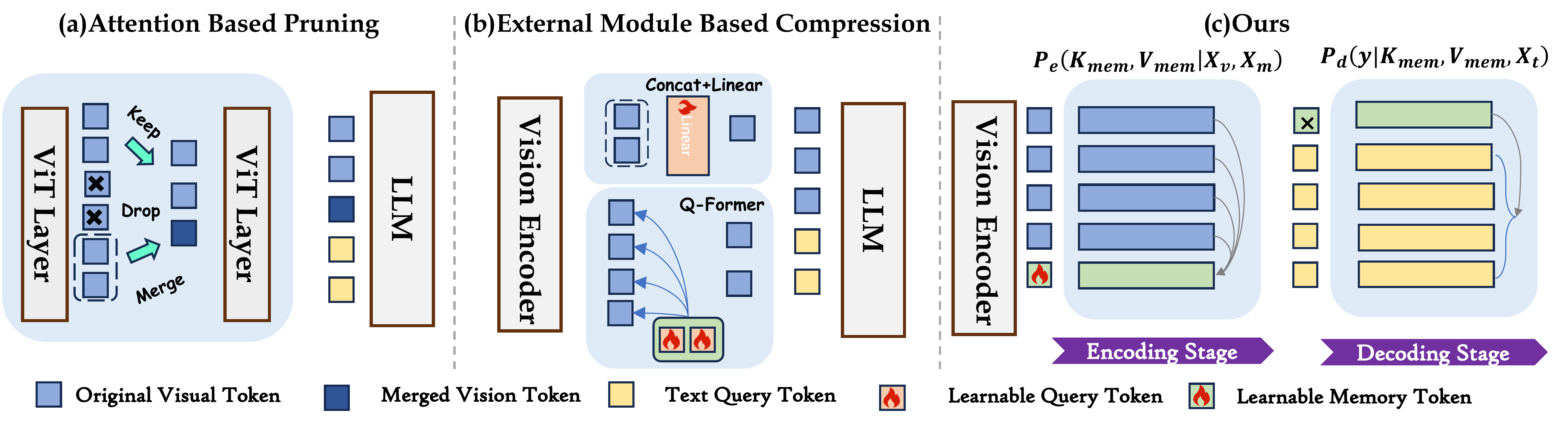}
    \caption{Comparison of Vision-Token Compression methods in the Vision Encoding Stage.}
    \label{fig1}
\end{figure*}
Large language models (LLMs)\cite{touvron2023llama,bai2023qwen,grattafiori2024llama} have demonstrated strong capabilities in natural language understanding and generation. Recently, by jointly modeling visual information and text, the powerful priors of LLMs have been successfully transferred to visual domains, giving rise to vision-language models (VLMs)\cite{liu2023visual,wang2024qwen2,alayrac2022flamingo,li2024llava} that can handle a wide range of complex tasks such as visual question answering and multimodal reasoning. However, high-resolution images are typically encoded into a large number of visual tokens, which dramatically increases the cost of self-attention computation and key–value (KV) cache storage\cite{kwon2023efficient}, thereby severely limiting the deployment of VLMs in resource-constrained and real-time scenarios\cite{yao2024minicpm}.

Prior studies have shown that visual inputs contain substantial redundancy~\cite{liang2022not,rao2021dynamicvit}. Inspired by such observations, numerous methods\cite{zhang2024sparsevlm,shang2025llava,chen2024image,yang2025visionzip,wen2025stop,zhang2025beyond,shao2025growing,alvar2025divprune,ye2025atp} have been proposed to compress visual tokens. Training-free methods typically perform heuristic compression on VLMs. Attention-based approaches represented by FastV\cite{chen2024image} and SparseVLM\cite{zhang2024sparsevlm} estimate token importance from text-to-vision attention in the LLM. However, as noted by~\cite{zhang2025beyond}, they are prone to attention shift, which can induce hallucinations~\cite{ji2023survey}. Another line of work performs similarity-based pruning inside the vision encoder. Methods such as DART~\cite{wen2025stop} and VisionZip~\cite{yang2025visionzip} remove tokens based on local feature similarity. While effective at eliminating local redundancy, they often fail to preserve text-guided global semantics~\cite{liao2025we}. As a result, both paradigms degrade sharply under high compression ratios. As shown in Fig.~\ref{fig0}, with only 9 tokens, SparseVLM loses details and global context, and its token selection is highly question-dependent. In contrast, our method captures both the overall scene and key regions with very few tokens, leading to the correct answer.
In contrast, training-based methods~\cite{li2023blip,chen2023minigpt,li2024inferenceoptimalvlmsneed,hu2024matryoshka,ye2025voco} improve robustness under tighter token budgets, but rely on additional compression modules or substantial retraining. Consequently, external-module-based methods often still provide limited gains under aggressive compression, while stronger methods typically come at the cost of much heavier training and reduced plug-and-play flexibility.

Visual token compression is essentially an encoding process that maps dense visual information into a compact token set while preserving essential semantics. This process hinges on strong information-encoding ability, which pretrained VLMs inherently possess. Therefore, rather than relying on heuristic pruning rules or external compression modules that require substantial retraining, we argue that better exploiting the priors of pretrained VLMs is key to effective visual token compression. In particular, given the strong priors already embedded in pretrained VLMs, an effective compression method should follow the native processing pattern of the VLM itself and achieve compression with only lightweight adaptation.

Motivated by this, we propose \textbf{VisCo}, an autoencoder that leverages priors of VLMs to achieve \textbf{Vi}sual token \textbf{s}elf-\textbf{Co}mpression during the vision encoding stage. Unlike prior approaches that either rely on attention and similarity for compression (Fig.~\ref{fig1}(a))\cite{yang2025visionzip,shang2025llava} or employ external modules with high retraining cost (Fig.~\ref{fig1}(b))\cite{chen2023minigpt,li2023blip}, VisCo reuses the VLM itself (Fig.~\ref{fig1}(c)) as the compressor: a small set of learnable memory tokens is introduced to interact with visual tokens in the encoder, while hierarchical visual information is passed from the encoder to the decoder to preserve task-relevant semantics under a reduced token budget. This design keeps the backbone intact and requires only lightweight adaptation. Our main contributions are summarized as follows:

\textbf{(1).} \textbf{An asymmetric intrinsic self-compression framework.} We propose VisCo, which uses the VLM itself to compress its own visual tokens with lightweight training while keeping the pretrained backbone unchanged.

\textbf{(2).} \textbf{A hierarchical information-passing mechanism.} We introduce a hierarchical mechanism to pass multi-granularity semantics from encoder layers to the decoder under compression.

\textbf{(3).} \textbf{Comprehensive evaluation and strong performance under extreme compression.} Extensive experiments on 3 VLM backbones and 6 benchmarks show that VisCo consistently outperforms existing methods, especially at high compression ratios. We further show that the learned compressed representations capture complementary information beyond compression alone.
\section{Related Work}
\subsection{Vision Language Models}
\begin{figure*}[t]
    \centering
    \includegraphics[width=\textwidth]{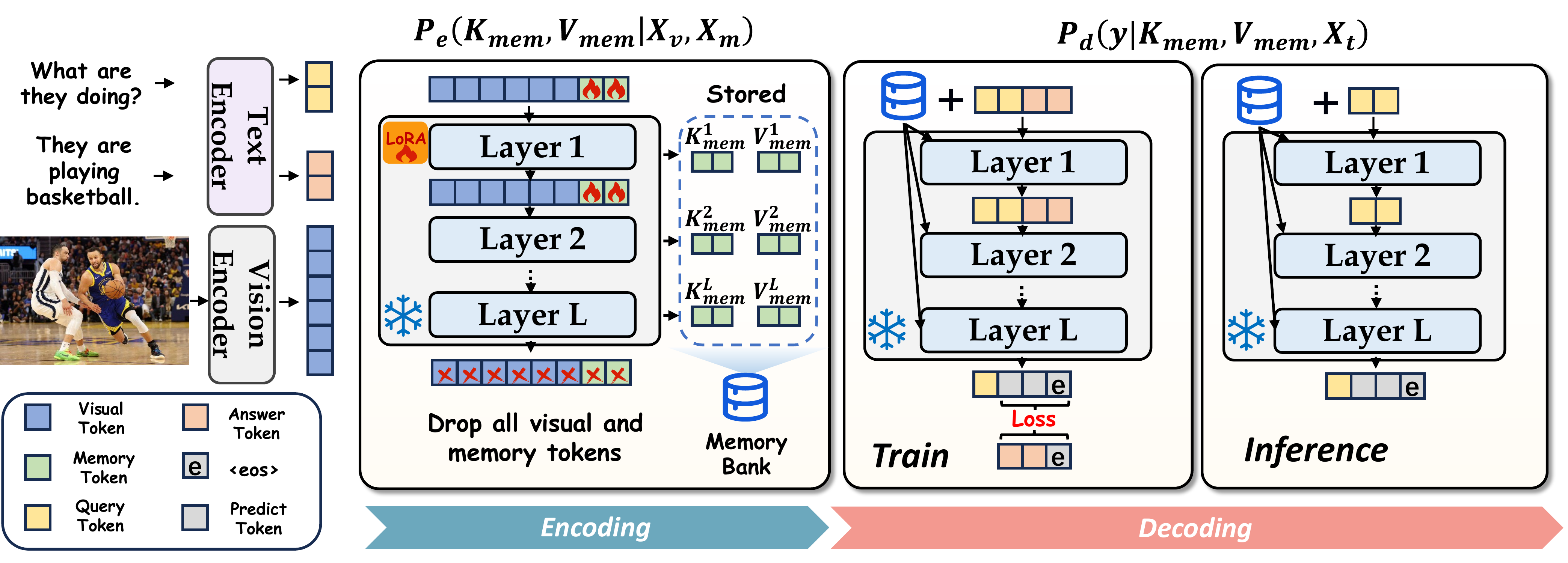}
    \caption{Illustration of the VisCo framework. During the encoding phase, hierarchical KV values corresponding to memory tokens are stored in a memory bank; during the decoding phase, these KV values are retrieved and hierarchically populated into the KV cache.}
    \label{fig2}
\end{figure*}
Recent advances in LLMs\cite{brown2020language,touvron2023llama,ouyang2022training} have driven the rapid development of VLMs, which typically consist of a vision encoder, a modality alignment module, and an LLM backbone. Representative VLMs, such as LLaVA\cite{liu2023visual}, BLIP-2\cite{li2023blip}, and InstructBLIP\cite{dai2023instructblip}, typically integrate pretrained visual encoders\cite{radford2021learning,zhai2023sigmoid} with LLM backbones via lightweight modality alignment modules, and many of these modular frameworks are trained under a two-stage paradigm that first establishes cross-modal alignment and subsequently equips the model with multimodal instruction-following capabilities. Despite their strong multimodal reasoning ability, current VLMs still struggle with fine-grained perception such as OCR and high-resolution detail understanding\cite{tong2024eyes}. To alleviate these limitations, recent efforts\cite{liu2024llavanext} have resorted to higher input resolutions to enhance visual perception and reduce hallucinations, consequently introducing substantially more visual tokens and imposing heavier computational overhead. To reduce the computational overhead, several recent VLMs\cite{wang2024qwen2,chen2024expanding,chen2024internvl,yao2024minicpm} have incorporated built-in visual compression mechanisms, including the dynamic-resolution design of Qwen2-VL\cite{wang2024qwen2} and the token reduction strategy adopted in InternVL\cite{chen2024internvl}. Yet the compression achieved by these architectures is still limited, which suggests that more dedicated visual token compression methods remain necessary for further improving efficiency without sacrificing fine-grained semantics or downstream reasoning performance.
\subsection{Token Compression Methods}
Training-free methods typically prune or merge visual tokens based on cross-modal attention scores or token similarity. Early approaches such as FastV\cite{chen2024image} and SparseVLM\cite{zhang2024sparsevlm} mainly perform attention-based token pruning, but as the token budget becomes tighter, they often suffer substantial performance degradation due to the loss of fine-grained details and global semantic information. To alleviate this issue, methods such as VisionZip\cite{yang2025visionzip} introduce token merging to preserve more global semantics during compression, yet their gains remain limited under high compression ratios.

Overall, training-free methods often remain competitive under mild compression, but such settings are not particularly challenging. Prior work\cite{liao2025we} has shown that, at low compression ratios, even simple input downsampling can outperform existing methods. Thus, the central challenge in visual token compression is to preserve semantic fidelity under aggressive compression.

To address this challenge, recent training-based methods can be roughly divided into two categories. One line of work, represented by PruMerge\cite{shang2025llava} and ACCM\cite{fu2025mitigating}, introduces lightweight compression modules with fine-tuning-level training cost. While efficient, such methods still provide only limited gains under aggressive compression. Another line of work adopts heavier training pipelines to improve performance under extreme token budgets. Earlier methods such as QueCC\cite{li2024inferenceoptimalvlmsneed} and Matryoshka Query\cite{hu2024matryoshka} rely on external compression modules or architectural modifications, leading to higher training cost while still requiring the VLM to adapt to a new compression mechanism. More recent concurrent works, such as VoCo-LLaMA\cite{ye2025voco} and C\&C\cite{bulat2025compress}, begin to explore compression within the VLM itself. However, they still rely on costly alignment, instruction tuning, or more specialized compression training, and therefore achieve strong performance at the cost of substantially modifying the original operating regime of the pretrained VLM. Different from both lines of work, VisCo pursues lightweight adaptation while directly leveraging the pretrained VLM itself for fine-grained hierarchical visual token self-compression. By following, rather than rewriting, the pretrained VLM’s native processing pattern, VisCo achieves performance competitive with heavyweight methods while retaining the flexibility and efficiency of lightweight adaptation.

\begin{table*}[t]
    \centering
    \caption{Comparison on LLaVA-1.5-7B under different token budgets.
    Avg. denotes the average percentage of performance maintained
    relative to the original LLaVA-1.5-7B. Methods marked with
    $\dagger$ are stronger references that are not directly comparable
    due to stronger training or different experimental settings.}
    \label{tab:llava}

    \begingroup
    \setlength{\tabcolsep}{3.5pt}
    \renewcommand{\arraystretch}{1.0}

    \begin{tabular*}{0.95\textwidth}{
        @{\extracolsep{\fill}}
        ccccccccc
        @{}
    }
        \specialrule{.08em}{0pt}{0pt}

        \textbf{Method}
        & \textbf{Token}
        & \textbf{GQA}
        & \textbf{MMB}
        & \textbf{MMB-CN}
        & \textbf{MME}
        & \textbf{POPE}
        & \textbf{MMVet}
        & \textbf{Avg.} \\

        \specialrule{.05em}{0pt}{0pt}

        LLaVA-1.5-7B
        & 576 & 62.0 & 64.3 & 58.3 & 1510.7 & 85.9 & 31.1
        & 100.0\% \\

        \specialrule{.05em}{0pt}{0pt}

        FastV {\scriptsize (ECCV24)}
        & 32 & 41.5 & 37.8 & 33.2 & 884.6 & 32.5 & 20.7
        & 57.6\% \\

        SparseVLM {\scriptsize (ICML25)}
        & 32 & 48.3 & 51.4 & 40.6 & 1046.7 & 67.9 & 18.6
        & 72.6\% \\

        PruMerge+ {\scriptsize (ICCV25)}
        & 32 & 51.1 & 56.8 & 47.0 & 940.8 & 70.9 & 21.4
        & 77.5\% \\

        DivPrune {\scriptsize (CVPR25)}
        & 32 & 54.9 & 57.6 & 49.1 & \textbf{1284.9}
        & 81.5 & 26.3 & 87.8\% \\

        VisPruner {\scriptsize (ICCV25)}
        & 32 & 52.2 & 58.4 & 52.7 & 1271.0
        & 72.7 & \textbf{28.8} & 87.8\% \\

        VisCo
        & 32 & \textbf{58.5} & \textbf{62.3} & \textbf{57.2}
        & 1152.9 & \textbf{81.9} & 27.9 & \textbf{91.8\%} \\

        {\color{lightgraytext}
        VoCo-LLaMA {\scriptsize (CVPR25)}}$\dagger$
        & {\color{lightgraytext}32}
        & {\color{lightgraytext}60.2}
        & {\color{lightgraytext}59.4}
        & {\color{lightgraytext}--}
        & {\color{lightgraytext}--}
        & {\color{lightgraytext}--}
        & {\color{lightgraytext}--}
        & {\color{lightgraytext}--} \\

        \specialrule{.05em}{0pt}{0pt}

        PruMerge+ {\scriptsize (ICCV25)}
        & 1 & 26.4 & 13.7 & 13.7 & 568.9 & 40.4 & 12.5
        & 35.4\% \\

        VisPruner {\scriptsize (ICCV25)}
        & 1 & 41.8 & 22.4 & 25.8 & 764.4 & 49.0 & 12.0
        & 48.8\% \\

        VisCo
        & 1 & \textbf{58.2} & \textbf{59.0} & \textbf{52.4}
        & \textbf{1191.2} & \textbf{78.2} & \textbf{20.7}
        & \textbf{85.3\%} \\

        {\color{lightgraytext}
        Matryoshka {\scriptsize (NIPS24)}}$\dagger$
        & {\color{lightgraytext}1}
        & {\color{lightgraytext}50.8}
        & {\color{lightgraytext}54.4}
        & {\color{lightgraytext}--}
        & {\color{lightgraytext}1144.0}
        & {\color{lightgraytext}74.5}
        & {\color{lightgraytext}--}
        & {\color{lightgraytext}--} \\

        {\color{lightgraytext}
        VoCo-LLaMA {\scriptsize (CVPR25)}}$\dagger$
        & {\color{lightgraytext}1}
        & {\color{lightgraytext}57.0}
        & {\color{lightgraytext}58.8}
        & {\color{lightgraytext}--}
        & {\color{lightgraytext}1323.3}
        & {\color{lightgraytext}81.4}
        & {\color{lightgraytext}--}
        & {\color{lightgraytext}--} \\

        \specialrule{.08em}{0pt}{0pt}
    \end{tabular*}

    \endgroup
\end{table*}
\section{Method}
We propose VisCo, a framework that compresses visual tokens by exploiting the priors of the VLM itself. As illustrated in Fig.~\ref{fig2}, VisCo is an asymmetric VLM autoencoder that introduces a lightweight trainable module in the encoding stage while keeping the entire VLM frozen during decoding.

\subsection{Overall Architecture: Asymmetric VLM Autoencoder with Shared Parameters}
VisCo is designed as a general solution for advanced VLMs such as Qwen2-VL\cite{wang2024qwen2} and LLaVA-1.5-7B\cite{liu2024improved}. Let the pretrained VLM backbone be denoted by $\Phi(\cdot)$. In the encoding stage, we augment the VLM with LoRA\cite{hu2022lora} adapter $\theta_{\mathrm{LoRA}}$ and introduce a small set of learnable memory tokens $X_m$. The encoder can then be written as $\Phi_E(\cdot; X_m, \theta_{\mathrm{LoRA}})$,
where the trainable parameters only include $X_m$ and $\theta_{\mathrm{LoRA}}$. 
In the decoding stage, we remove the LoRA adapters used in the encoder and directly reuse the shared pretrained LLM backbone as the decoder. This asymmetric design adapts only the encoder while keeping the decoder frozen, so that VisCo requires only lightweight adaptation of the pretrained VLM’s existing capabilities to the compression setting rather than relearning the whole multimodal generation process from scratch, which is different from methods like VoCo-LLaMA\cite{ye2025voco}.
\subsection{LLM-based Encoding Process}
\subsubsection{Sequence Construction under Causal Masking}
Following \cite{liu2023visual}, given an image $I^{\mathrm{origin}}$, we first utilize the function $f_{\mathrm{resize}}$ that
rescales the image to a target resolution to obtain $I$. The vision encoder $f_v$ and the multimodal projector $g$ encode the origin $I$ into a sequence of visual tokens $X_v = g\big(f_v(I)\big) \in \mathbb{R}^{N_v \times D}$ where $N_v$ is the number of visual tokens and $D$ is the feature dimension.
We then append $N_m$ learnable memory tokens $X_m \in \mathbb{R}^{N_m \times D}$, where $N_m \ll N_v$, and form the joint input sequence
$X = [X_v; X_m] \in \mathbb{R}^{(N_v+N_m)\times D}$. VisCo strictly follows the standard causal masking mechanism used in decoder-only VLMs, where a token at position $i$ is only allowed to attend to tokens at positions $j \le i$.
Therefore, by placing $X_m$ after the visual tokens $X_v$ in the sequence, each memory token can naturally attend to all visual tokens under the native attention pattern of the pretrained backbone. In this way, VisCo does not introduce any customized interaction rule, but instead simply leverages the model’s intrinsic attention prior to let the memory tokens aggregate rich visual information from the visual tokens.

\subsubsection{Hierarchical Information Aggregation}

Transformer-based LLMs are known to process information in a
hierarchical manner. Prior works have shown that Transformer language
models exhibit layer-wise functional specialization: lower layers tend
to encode surface and local syntactic patterns, middle layers focus on
syntactic structure, while higher layers capture semantics and
discourse-level phenomena~\cite{rogers2020primer}. A compact representation for visual compression should preserve not only the final semantic summary, but also the hierarchical structure of visual information. Relying only on the final encoder output would collapse this hierarchy and make it harder for the decoder to recover fine-grained details under aggressive compression.
Instead of passing the output of the encoder to the decoder, which destroys the layer-wise structure of compression information,
VisCo is motivated by the structural similarity between the encoder and decoder. It adopts a hierarchical compression scheme that transfers hierarchical information from the encoder to the decoder, leveraging the self-attention mechanism of LLMs. For each layer $l = 1, \ldots, L$, let $K^{(l)}$ and $V^{(l)}$ denote the key and value matrices of the self-attention computed over the joint sequence $X = [X_v; X_m]$. We construct a layer-wise memory bank by retaining only the key and value entries corresponding to the memory tokens:
\begin{equation}
\begin{gathered}
K^{(l)}_{\mathrm{mem}} = K^{(l)}[N_v+1 : N_v+N_m], \\
V^{(l)}_{\mathrm{mem}} = V^{(l)}[N_v+1 : N_v+N_m].
\end{gathered}
\end{equation}
We define $K_{\mathrm{mem}}=\{K^{(l)}_{\mathrm{mem}}\}_{l=1}^{L}$ and $V_{\mathrm{mem}}=\{V^{(l)}_{\mathrm{mem}}\}_{l=1}^{L}$, and model the encoding stage as:
\begin{equation}
  P_e\!\left(K_{\mathrm{mem}}, V_{\mathrm{mem}} \mid X_v, X_m; \Phi_{\mathrm{E}}\right).
  \label{eq:VisCo-enc}
\end{equation}
Under causal masking, each memory token can attend to all preceding visual tokens.
Consequently, shallow layers of $\{K^{(l)}_{\mathrm{mem}}, V^{(l)}_{\mathrm{mem}}\}$ capture low-level cues
such as textures and local patterns, whereas deeper layers encode progressively
more abstract semantic information. In practice, we collect the KV pairs
associated with memory tokens at every layer into the memory bank, discard the
final encoder hidden states, and rely solely on these layer-wise memory KV
caches during decoding.

\subsection{Hierarchical Prefix Decoding}
\label{sec:hier-prefix-decoding}
\begin{table*}[t]
    \centering
    \caption{Performance comparison of different token compression methods on Qwen2-VL-2B and Qwen2-VL-7B. Here, Avg represents the average percentage of performance maintained. $\ddag$ indicates additional fine-tuning.}
    \label{tab: Qwen}

    \begingroup
    \setlength{\tabcolsep}{1.5pt}
    \setlength{\extrarowheight}{0.8pt}
    \setlength{\arrayrulewidth}{.05em}
    \setlength{\aboverulesep}{0pt}
    \setlength{\belowrulesep}{0pt}
    \setlength{\cmidrulesep}{0pt}
    
    \begin{tabular*}{\textwidth}{
    @{}
    >{\columncolor{tablegray}\centering\arraybackslash}m{2.2cm}|
    >{\centering\arraybackslash}m{2.7cm}|
    @{\extracolsep{\fill}}
    *{7}{c}
    @{}
}
        \specialrule{.08em}{0pt}{0pt}

        \cellcolor{white}\textbf{Base Model}
        & \textbf{Method}
        & \textbf{GQA}
        & \textbf{MMB}
        & \textbf{MMB-CN}
        & \textbf{MME}
        & \textbf{POPE}
        & \textbf{MMVet}
        & \textbf{Avg.} \\

        \specialrule{.05em}{0pt}{0pt}

        \rowcolor{tablegray}
        & \multicolumn{8}{c}{Upper Bound, 144 Tokens (100\%)} \\

        & Origin
        & 59.8 & 67.3 & 61.8 & 1465.8 & 81.5 & 42.5
        & \textbf{100.0\%} \\

       \cmidrule[.05em]{2-9}

        \rowcolor{tablegray}
        & \multicolumn{8}{c}{
            Retain 36 Tokens
            (\textcolor{green!60!black}{$\downarrow$ 75.0\%})
        } \\

        & FastV {\scriptsize (ECCV24)}
        & 49.5 & 62.2 & 59.3 & 1289.3 & 68.9 & 26.2
        & 84.2\% \\

        & PruMerge+ {\scriptsize (ICCV25)}
        & 52.1 & 49.3 & 46.8 & \textbf{1387.2} & 70.2 & 28.4
        & 80.6\% \\

        & VisionZip$\ddag$ {\scriptsize (CVPR25)}
        & 53.2 & \textbf{64.3} & 56.7 & 1372.4 & 75.3 & 35.5
        & 91.0\% \\

        & VisCo
        & \textbf{61.1} & 64.0 & \textbf{58.9} & 1368.3
        & \textbf{82.9} & \textbf{35.6}
        & \textbf{95.2\%} \\

       \cmidrule[.05em]{2-9}
    
        \rowcolor{tablegray}
        & \multicolumn{8}{c}{
            Retain 18 Tokens
            (\textcolor{green!60!black}{$\downarrow$ 87.5\%})
        } \\

        & FastV {\scriptsize (ECCV24)}
        & 46.2 & 56.0 & 47.9 & 1159.2 & 49.0 & 18.1
        & 70.0\% \\

        & PruMerge+ {\scriptsize (ICCV25)}
        & 46.6 & 47.9 & 26.5 & 1190.7 & 61.0 & 18.1
        & 65.1\% \\

        & VisionZip$\ddag$ {\scriptsize (CVPR25)}
        & 48.5 & 55.2 & 41.6 & 1300.9 & 62.7 & 25.8
        & 76.1\% \\

        \multirow[c]{-12}{*}{\textbf{Qwen2-VL-2B}}
        & VisCo
        & \textbf{59.7} & \textbf{62.9} & \textbf{57.8}
        & \textbf{1358.0} & \textbf{82.4} & \textbf{28.9}
        & \textbf{91.4\%} \\

        \specialrule{.05em}{0pt}{0pt}

        \rowcolor{tablegray}
        & \multicolumn{8}{c}{Upper Bound, 144 Tokens (100\%)} \\

        & Origin
        & 64.8 & 76.1 & 71.6 & 1664.8 & 82.6 & 56.1
        & \textbf{100.0\%} \\

       \cmidrule[.05em]{2-9}

        \rowcolor{tablegray}
        & \multicolumn{8}{c}{
            Retain 36 Tokens
            (\textcolor{green!60!black}{$\downarrow$ 75.0\%})
        } \\

        & FastV {\scriptsize (ECCV24)}
        & 55.4 & 70.7 & 64.6 & 1492.0 & 65.8 & 36.6
        & 83.9\% \\

        & PruMerge+ {\scriptsize (ICCV25)}
        & 59.2 & 71.8 & 63.1 & 1518.7 & 73.9 & 37.5
        & 86.9\% \\

        & VisionZip$\ddag$ {\scriptsize (CVPR25)}
        & 58.3 & 72.4 & 63.4 & 1530.0 & 75.3 & 39.1
        & 87.7\% \\

        & VisCo
        & \textbf{62.6} & \textbf{74.0} & \textbf{70.0}
        & \textbf{1551.0} & \textbf{83.3} & \textbf{40.4}
        & \textbf{92.9\%} \\

       \cmidrule[.05em]{2-9}

        \rowcolor{tablegray}
        & \multicolumn{8}{c}{
            Retain 18 Tokens
            (\textcolor{green!60!black}{$\downarrow$ 87.5\%})
        } \\

        & FastV {\scriptsize (ECCV24)}
        & 52.8 & 65.4 & 59.4 & 1395.2 & 60.7 & 33.4
        & 77.9\% \\

        & PruMerge+ {\scriptsize (ICCV25)}
        & 52.8 & 61.2 & 47.7 & 1351.7 & 63.7 & 30.6
        & 73.6\% \\

        & VisionZip$\ddag$ {\scriptsize (CVPR25)}
        & 55.3 & 66.4 & 58.3 & 1443.8 & 71.7 & 33.9
        & 81.3\% \\

        \multirow[c]{-12}{*}{\textbf{Qwen2-VL-7B}}
        & VisCo
        & \textbf{61.8} & \textbf{70.0} & \textbf{68.8}
        & \textbf{1496.5} & \textbf{81.7} & \textbf{39.2}
        & \textbf{90.4\%} \\

        \specialrule{.08em}{0pt}{0pt}
    \end{tabular*}

    \endgroup
\end{table*}
Our encoder and decoder share the same backbone parameters. Specifically, the
decoding stage reuses the pretrained VLM $\Phi$ with all LoRA adapters
removed. To enable the decoder to exploit
the hierarchical memory information collected during encoding, we directly
populate the decoder's KV cache with the layer-wise memory bank
$\{K^{(l)}_{\mathrm{mem}}, V^{(l)}_{\mathrm{mem}}\}_{l=1}^{L}$ and then
perform autoregressive generation on top of this cache. Accordingly, we model the decoding stage as
\begin{equation}
P_d\!\left(y \mid K_{\mathrm{mem}}, V_{\mathrm{mem}}, X_t; \Phi\right),
  \label{eq:VisCo-dec}
\end{equation}
where $y$ denotes the answer sequence and $X_t \in \mathbb{R}^{N_t \times D}$ denotes the embedding sequence of textual prompt with length $N_t$.

Benefiting from the strong priors of the underlying VLM, we do not introduce any additional pre-training objectives.
Instead, we fine-tune VisCo end-to-end with teacher forcing, which keeps the training cost low.
We maximize the conditional likelihood of the ground-truth answer given the compressed memory and textual inputs. The resulting loss is
\begin{equation}
\mathcal{L}_{\mathrm{FT}}
= - \sum_{i=1}^{N_a} 
    \log p_\theta\big(a_i \mid  K_{\mathrm{mem}}, V_{\mathrm{mem}}, X_t, A_{<i}\big),
\end{equation}
where $A = \{a_1,\ldots,a_{N_a}\}$ denotes the answer token sequence with length $N_a$ and $A_{<i}$ represents answer tokens located before the current predicted token $a_i$. 
$\theta$ represents trainable parameters, consisting of $X_m$ and $\theta_{\mathrm{LoRA}}$.
Unlike Prefix-Tuning\cite{li2021prefix}, which uses $L$ per-layer MLPs to project prefix tokens into key/value prefixes, we directly reuse the encoder-side memory KV caches as hierarchical prefixes. Since the encoder-side KV caches already carry layer-wise, fine-grained information and the encoder and decoder share the same representational and attention priors via parameter sharing, the decoder can directly reuse them without any projection.

\section{Experiments}
In this section, we comprehensively evaluate the performance and efficiency of our solution using 6 benchmarks across 3 VLM models.
\subsection{Implementation Details}
We fine-tune VisCo for one epoch on a $\sim$10\% subset of LLaVA-665K \cite{liu2024improved}, consisting of the first turn of each conversation. We apply LoRA to the Q/V projections with $rank=64$ and $\alpha=128$, using a learning rate of $5\times10^{-5}$. We evaluate VisCo on LLaVA-1.5-7B and Qwen2-VL (2B/7B). Notably, Qwen2-VL incorporates native $4\times$ compression, serving as a rigorous testbed to validate VisCo's robustness on already-compact visual representations. 

\begin{figure}[htbp]
\centering
\includegraphics[width=0.9\columnwidth]{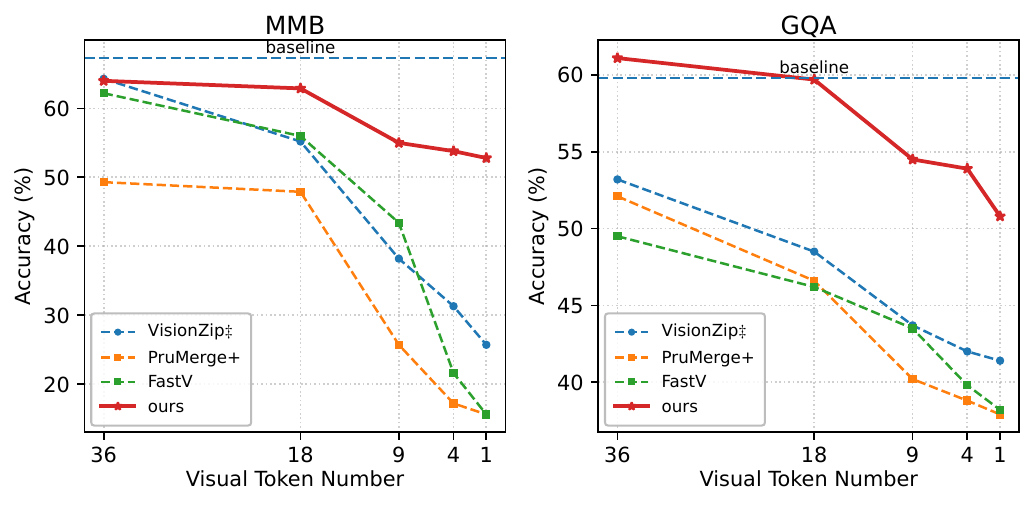}
\caption{Ablation Study of Token Compression Rates on MMB and GQA.}
\label{fig3}
\end{figure}

\begin{figure*}[htbp]
    \centering
    \includegraphics[width=\textwidth]{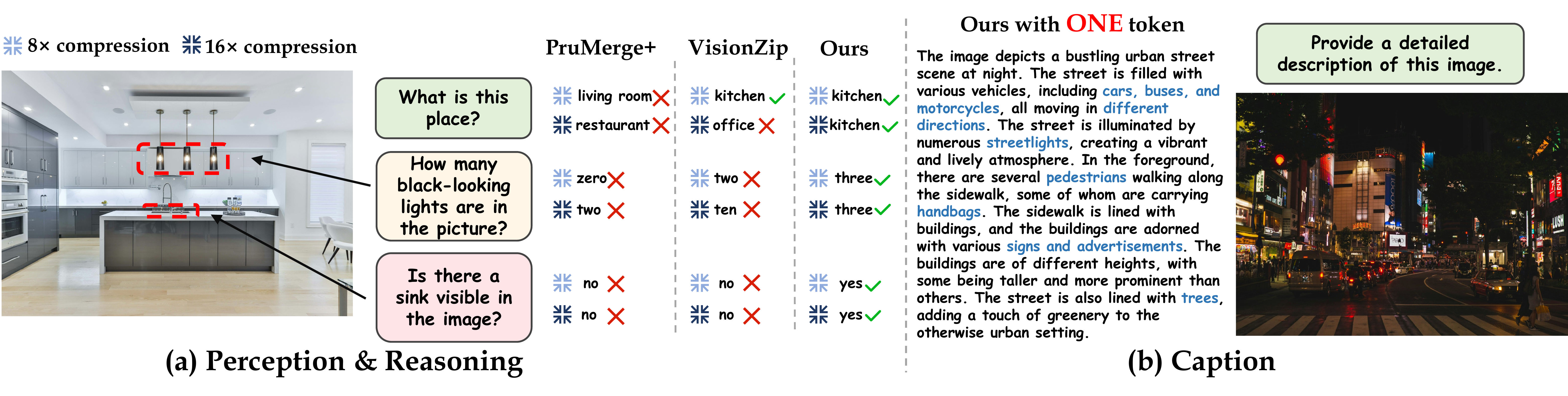}
    \caption{Qualitative evaluation under high compression ratio. (a) Perception and reasoning comparison of existing compression methods versus VisCo. (b) Caption generation from a single VisCo token. All experiments were conducted on Qwen2-VL-2B}
    \label{fig4}
\end{figure*}
\subsection{Datasets}
We evaluated VisCo across six multimodal benchmarks, including MME\cite{Fu2023MMEAC}, MMB\cite{liu2024mmbench}, MMB-CN, MMVet\cite{yu2023mm}, GQA\cite{hudson2019gqa}, and POPE~\cite{li2023evaluating}. We report standard metrics for each benchmark, including the perception score on MME, accuracy on MMB/MMB-CN and GQA, the GPT-4\cite{achiam2023gpt} assisted score on MMVet, and F1 on POPE. We also report Avg, defined as the unweighted mean of the retained-score ratio (compressed/original) across benchmarks. 
\subsection{Baselines}
We compare VisCo against six approaches. FastV\cite{chen2024image}, SparseVLM\cite{zhang2024sparsevlm}, DivPrune\cite{alvar2025divprune}, and VisPruner\cite{zhang2025beyond} are plug-and-play methods that can be applied directly without extra fine-tuning. Also, we include two fine-tuned baselines: PruMerge+\cite{shang2025llava} fine-tuned for 1 epoch on the full LLaVA-665K, and VisionZip$\ddag$\cite{yang2025visionzip} fine-tuned for 1 epoch on the same 10\% subset of LLaVA-665K as VisCo. $\ddag$ denotes VisionZip with additional fine-tuning. In addition, we further include two stronger references, VoCo-LLaMA\cite{ye2025voco} and MatryoshkaQuery\cite{hu2024matryoshka}. Among all VLM token compression methods, VoCo-LLaMA represents a SOTA reference. However, both methods require substantially retraining of the VLM, including re-alignment and large-scale instruction tuning, and therefore are not directly comparable to our lightweight fine-tuning setting. We report both methods for reference only. Due to differences in publicly available implementations and backbone compatibility across methods, the set of comparable baselines varies slightly across VLM backbones.

\begin{table}[htbp]
    \centering
    \caption{Ablation Study Comparing Value-Passing and Hierarchical KV-Pass for Encoder–Decoder Information Transfer. Score is the retained performance percentage.}
    \label{tab: Hiera}    
    \resizebox{0.95\columnwidth}{!}{%
    \setlength{\tabcolsep}{4pt}%
    \renewcommand{\arraystretch}{1.05}%
    \setlength{\aboverulesep}{0pt}%
    \setlength{\belowrulesep}{0pt}%
    \begin{tabular}{l|cc|cc|cc}
        \toprule
        \multirow{2}{*}{\textbf{Method}} &
        \multicolumn{2}{c|}{\textbf{GQA}} &
        \multicolumn{2}{c|}{\textbf{MMB}} &
        \multicolumn{2}{c}{\textbf{POPE}} \\
        & Acc. & Score & Acc. & Score & F1. & Score  \\
        \midrule
        \rowcolor{gray!30}
        \multicolumn{7}{c}{Upper Bound, All 144 Tokens (\textbf{100\%})} \\
        Origin &
        59.8 & 100\% &
        67.3 & 100\% &
        81.5 & 100\%  \\

        \rowcolor{gray!30}
        \multicolumn{7}{c}{Retain 36 Tokens (\textcolor{green!60!black}{$\downarrow$ 75.0\%})} \\
        Our-ValuePass &
        51.8 & 86.6\% &
        59.5 & 88.4\% &
        79.6 & 97.7\%  \\
        Our-HierKVPass &
        61.1 & 102.2\% &
        64.0 & 95.1\% &
        82.9 & 101.7\%  \\

        \rowcolor{gray!30}
        \multicolumn{7}{c}{Retain 18 Tokens (\textcolor{green!60!black}{$\downarrow$ 87.5\%}) } \\
        Our-ValuePass &
        49.6 & 82.9\% &
        54.4 & 80.8\% &
        77.6 & 95.2\% \\
        Our-HierKVPass  &
        59.7 & 99.8\% &
        62.9 & 93.5\% &
        82.4 & 101.1\%  \\

        \rowcolor{gray!30}
        \multicolumn{7}{c}{Retain 2 Tokens (\textcolor{green!60!black}{$\downarrow$ 98.6\%})} \\
        Our-ValuePass &
        45.8 & 76.6\% &
        50.1 & 74.4\% &
        74.9 & 91.9\%  \\
        Our-HierKVPass &
        50.8 & 84.9\% &
        53.3 & 79.2\% &
        79.7 & 97.8\% \\
        \bottomrule
    \end{tabular}%
    } 
\end{table}

\subsection{Main Results}
\noindent \textbf{Results on LLaVA-1.5.} As shown in Table~\ref{tab:llava}, we first evaluate VisCo on the classic LLaVA-1.5-7B under high-compression settings. When compressing the original 576 visual tokens to 32, VisCo still preserves 91.8\% of the original performance, outperforming all directly comparable baselines and surpassing VisPruner by 4.0 points in Avg. It also achieves the best results on four benchmarks, indicating that VisCo can retain task-relevant visual information more effectively under aggressive compression. When the token budget is further reduced to a single token, the advantage of VisCo becomes even more pronounced. Despite this extreme compression ratio, VisCo still retains 85.3\% of the original performance, exceeding PruMerge+ and VisPruner by 49.9 and 36.5 points in Avg., respectively. This suggests that even a very small number of memory tokens can still form compact yet effective visual representations.

For further reference, we also include two stronger training-based methods, VoCo-LLaMA and MatryoshkaQuery. Although they require substantially more large-scale retraining and are therefore not comparable to our lightweight fine-tuning setting, VisCo still consistently outperforms MatryoshkaQuery and even surpasses VoCo-LLaMA on GQA and MMB under the 1-token setting, further demonstrating its strong competitiveness in high-compression scenarios.

\noindent \textbf{Results on Qwen2-VL.} To assess generality, we further benchmark VisCo using Qwen2-VL. As shown in Table~\ref{tab: Qwen}, on Qwen2-VL-2B, VisCo retains $95.2\%$ of the full-token performance when compressing to 36 tokens, exceeding VisionZip$\ddag$ by $4.2\%$. When the budget is tightened to 18 tokens, prior methods degrade substantially: even the strongest baseline on Qwen2-VL, VisionZip$\ddag$, retains only $76.1\%$, whereas VisCo still maintains $91.4\%$. On Qwen2-VL-7B, VisCo remains consistently competitive under an $8\times$ compression ratio and outperforms VisionZip$\ddag$ by $9.1\%$.

\noindent \textbf{Analysis.} VisCo consistently outperforms directly comparable baselines and remains competitive with stronger references. Its advantage is especially evident on Qwen2-VL, which already incorporates a built-in visual compression mechanism. When handling denser visual information, existing compression methods suffer severe performance degradation, whereas VisCo remains remarkably stable. These results further validate the rationality of exploiting the intrinsic capabilities of LLMs for compression. Moreover, the gains are unlikely to come from fine-tuning or answer memorization. PruMerge+ and VisionZip$\ddag$  are trained on LLaVA-665K using the full set or a 10\% subset, yet they still lag behind VisCo. This suggests that VisCo benefits primarily from its intrinsic memory-based compression rather than supervision. Notably, despite being trained without Chinese data, VisCo still achieves the best performance on MMB-CN, further supporting the effectiveness of our mechanism.

\subsection{Ablation Study}
\subsubsection{Ablation Study of Reduction Ratios}
To explore VisCo's compression limits, we compare different compression strategies on Qwen2-VL-2B across a range of compression ratios on GQA and MMB. As shown in Fig.~\ref{fig3}, the performance of existing methods drops sharply as the number of retained visual tokens decreases, and all of them converge to a similar low accuracy when only a single token is preserved. In contrast, our method still achieves accuracies of 50.8 and 52.8 on GQA and MMB, respectively, outperforming all competing methods even when they keep 9 tokens. 

\begin{table}[t]
\centering
\caption{Comparison of VisCo+, the original model, and Direct SFT on 6 benchmarks with Qwen2-VL-2B. Direct SFT denotes directly fine-tuning the original model using the same training configuration as VisCo+. The best results are in bold and the second-best results are underlined.}
\label{tab:mm_compare}
\resizebox{0.95\columnwidth}{!}{
\begin{tabular}{lcccccc}
\toprule
Method & MMB & MMB-CN & GQA & MME & POPE & MMVet \\
\midrule
Origin        & \underline{67.3} & 61.8 & \underline{59.8} & \underline{1465.8} & 81.5 & \textbf{42.5} \\
Direct SFT            & 66.4 & \underline{62.2} & 57.1 & 1354.5 & \underline{82.6} & \underline{40.3} \\
\textbf{VisCo+} & \textbf{69.6} & \textbf{67.8} & \textbf{59.9} & \textbf{1491.2} & \textbf{84.6} & 37.2 \\
\bottomrule
\end{tabular}
}
\end{table}

 As shown in Fig.~\ref{fig4} (a), we compare VisCo with existing approaches under both $8\times$ and $16\times$ compression ratios. At high compression ratios, conventional methods miss fine-grained details, causing failures on complex reasoning and even hallucinations on simple scene classification due to weakened global semantics. In contrast, VisCo consistently answers all perception and reasoning tasks correctly. Fig.~\ref{fig4} (b) further showcases the captioning capability of VisCo. Remarkably, with only a single token, VisCo is able to generate a caption containing both detailed object-level descriptions and coherent global semantics. 
\begin{figure}[htbp]
\centering
\includegraphics[width=0.9\columnwidth]{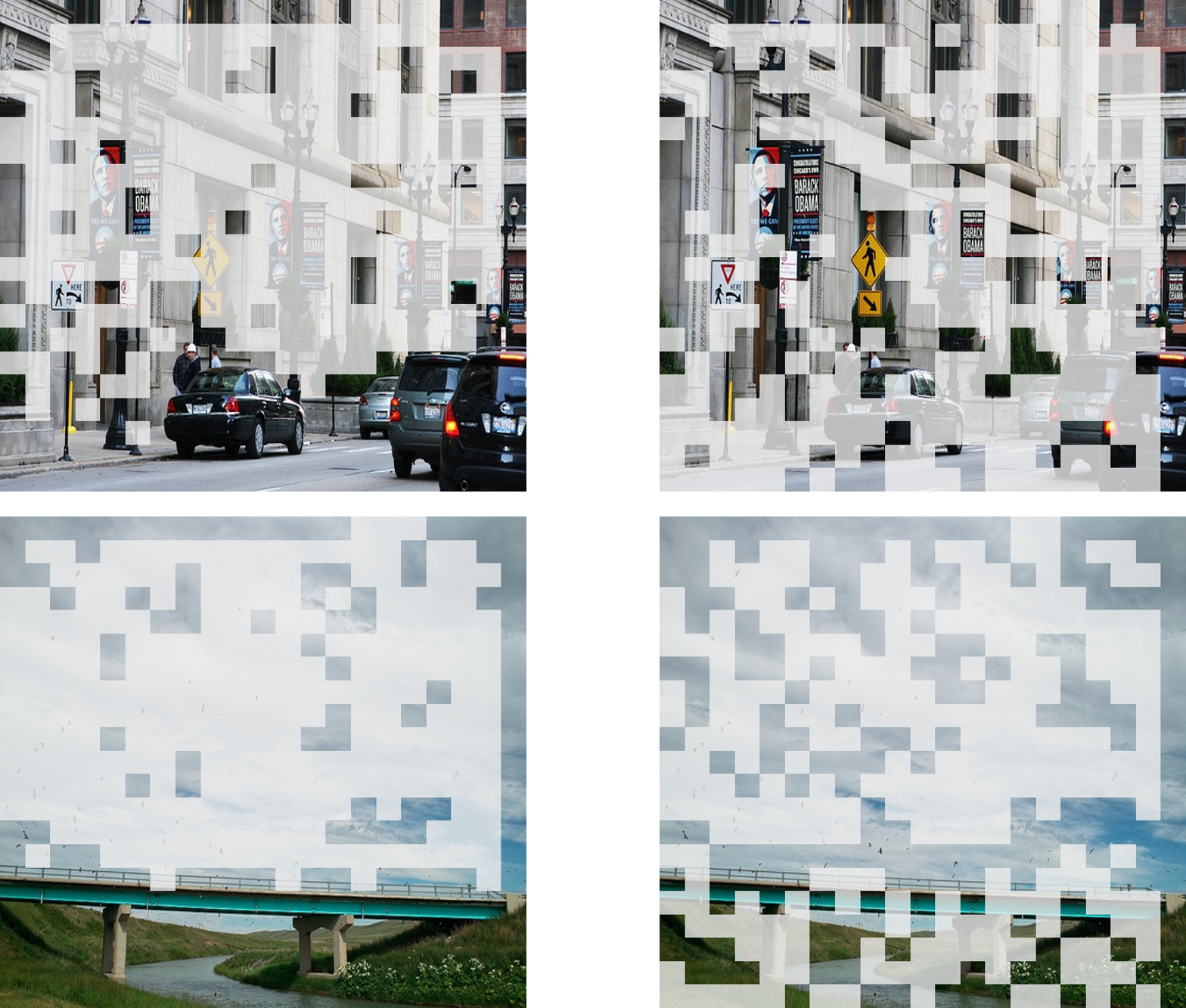}
\caption{Visualization of attention in the encoding stage of Qwen2-VL-2B with 36 memory tokens.
The transparent regions on the left highlight the top-175 visual tokens attended by the first four memory tokens, while those on the right correspond to the last four memory tokens.}
\label{fig5}
\end{figure}
 
\subsubsection{Ablation Study of Hierarchical Aggregation and Decoding}
To validate the effectiveness of our hierarchical KV-cache propagation scheme for compression, we implemented an alternative method: a value-passing variant that directly feeds the outputs of the encoder into the decoder. We conducted experiments on Qwen2-VL-2B under three different compression ratios using GQA, MMB and POPE. As shown in Table~\ref{tab: Hiera}, the results demonstrate that Our-HierKVPass consistently outperforms Our-ValuePass, confirming that hierarchical information propagation indeed substantially improves model performance.
Furthermore, we observe that Our-ValuePass still surpasses the other baseline methods listed in Table~\ref{tab: Qwen}, especially at high compression rates. This suggests that the performance gains do not arise solely from hierarchical aggregation, but also from the autoencoder architecture itself, which is able to better exploit the strong priors of the LLM.
\subsubsection{Ablation Study of Memory Tokens as Complementary Representations}
As shown in Table~\ref{tab: Qwen}, we observe an interesting phenomenon in the main experiments. On benchmarks such as GQA and POPE, models using compressed tokens can even outperform the original model. This suggests that memory tokens are not merely a subset or summary of the original visual tokens. Instead, they provide complementary representations and encode visual information from a new perspective.

To verify this hypothesis, we propose VisCo+. It is built upon the VisCo model trained with 36 retained tokens in Table~\ref{tab: Qwen}. During decoding, it jointly leverages the key-value pairs of the original visual tokens and the memory tokens. As shown in Table~\ref{tab:mm_compare}, we compare VisCo+, the original model, and a directly fine-tuned variant on Qwen2-VL-2B. The results show that VisCo+ improves over the original model on five of the six
benchmarks. The comparison with direct SFT further shows that the gain does not come from additional fine-tuning. These results support our claim that memory tokens are not simple compressed representations. They instead act as complementary representations that enrich the original visual features.

\begin{table}[t]
\centering
\caption{Ablation study of dropping different ranges of memory tokens on Qwen2-VL-2B. The best results are in bold and the second-best results are underlined.}
\label{tab: Drop}
\resizebox{\columnwidth}{!}{
\begin{tabular}{lccccc}
\toprule
Method & MMB & MMB-CN & GQA & MME & POPE \\
\midrule
36token         & \textbf{64.0} & \textbf{58.9} & \textbf{61.1} & \textbf{1368.3} & \textbf{82.9} \\
drop(0,17)      & 55.7          & 52.9          & 56.8          & 1305.6          & 76.2          \\
drop(18,35)     & \underline{60.0} & \underline{56.2} & \underline{57.0} & \underline{1337.4} & \underline{81.6} \\
\bottomrule
\end{tabular}
}
\end{table}
\subsubsection{How Do Memory Tokens Work?}
To investigate how memory tokens transfer information between the encoder and decoder, and to assess the importance of memory tokens at different positions, we first conduct an ablation study by dropping different ranges of memory tokens. As shown in Table~\ref{tab: Drop}, the standard VisCo uses 36 memory tokens during inference. Dropping the first 18 tokens causes a substantial performance decline, while dropping the last 18 tokens has a much smaller effect. This trend is consistent across five benchmarks.

This observation raises a natural question: are the later memory tokens redundant? To answer this, we further visualize the attention from memory tokens at different positions to the original visual tokens. As shown in Fig.~\ref{fig5}, the first four memory tokens attend more densely to the main objects in the image, whereas the last four tokens exhibit much sparser attention and tend to focus on background regions and local details. This finding also explains the results in Table~\ref{tab: Drop}. Early memory tokens are more likely to encode the main content of the image, and most benchmark questions are primarily centered on the main objects. Therefore, these tokens are more critical for performance. However, this does not mean that the later tokens are redundant. Instead, they capture scattered details and background information, which helps enrich the model’s overall scene perception.
\begin{table}[t]
    \centering
    \caption{Efficiency comparison under different token budgets on LLaVA-1.5-7B.}
    \label{tab:efficiency}
     \small
    \renewcommand{\arraystretch}{0.8}
    \setlength{\tabcolsep}{2.2pt}
    \resizebox{\columnwidth}{!}{
    \begin{tabular}{cccccc}
        \toprule
        Token & Method & \makecell[c]{Compression\\Time (ms)} & \makecell[c]{Decode Time\\(ms/token)} & \makecell[c]{Inference\\Time (s)} & \makecell[c]{KV\\Cache (MB)} \\
        \midrule
        576 & Origin    & --   & 21.2 & 5.5 & 288.0 \\
        \midrule
        \multirow{3}{*}{128}
            & FastV     & --   & 20.3 & 5.2 & 78.8 \\
            & VisionZip$\ddag$  & --   & 19.6 & 5.0 & 72.0 \\
            & VisCo     & 68.9 & 19.6 & 5.0 & 72.0 \\
        \midrule
        \multirow{3}{*}{32}
            & FastV     & --   & 19.4 & 5.0 & 26.4 \\
            & VisionZip$\ddag$  & --   & 18.6 & 4.8 & 18.0 \\
            & VisCo     & 63.7 & 18.6 & 4.8 & 18.0 \\
        \bottomrule
    \end{tabular}
    }
\end{table}

\begin{figure}[htbp]
\centering
\includegraphics[width=\columnwidth]{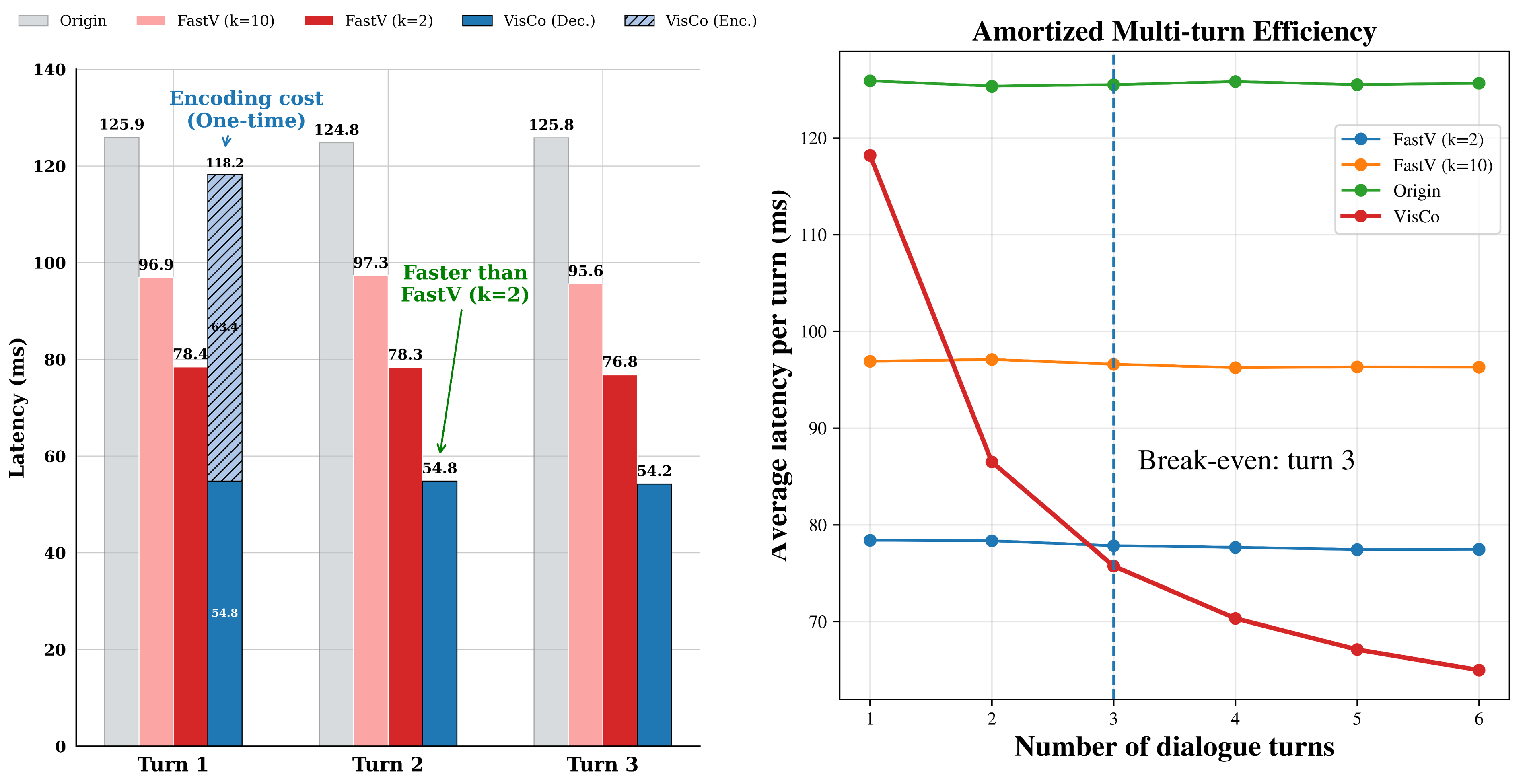}
\caption{Multi-turn efficiency on MME with LLaVA-1.5-7B. Left: per-turn latency. Right: amortized average latency per turn over dialogue turns, with VisCo reaching break-even at turn 3.}
\label{fig6}
\end{figure}
\subsection{Efficiency Analysis}
To evaluate the efficiency of VisCo, we conduct analyses on both long-response and short-response scenarios using MMVet\cite{yu2023mm} and MME\cite{Fu2023MMEAC}, respectively, on a single 48GB NVIDIA RTX 6000 GPU.

On MMVet, we fix the output length of LLaVA-1.5-7B to 256 tokens and compare VisCo with FastV and VisionZip$\ddag$. As shown in Table~\ref{tab:efficiency}, although VisCo introduces extra compression overhead due to its additional LLM-based encoding step, it achieves the same decoding speed as VisionZip and consistently outperforms Origin and FastV under both the 128-token and 32-token settings. In long-response scenarios, the decoding acceleration offsets the compression overhead, making VisCo 0.2 s faster than FastV in both settings. VisCo also significantly reduces KV-cache storage.

On MME, we evaluate efficiency in a short-answer setting, where the compression overhead becomes more noticeable. However, in practical applications, users often ask multiple questions about the same image. For methods such as FastV and SparseVLM, which rely on the current question text to compress visual tokens, the visual tokens must be compressed again for each query. In contrast, VisCo compresses the image once and reuses the cached representations across dialogue turns. As shown in Fig.~\ref{fig6}, although VisCo already achieves lower latency than Origin in the first round of question answering, its response time is still noticeably slower than that of FastV due to the one-time compression overhead. As the number of dialogue turns increases, however, the advantage of cache reuse in VisCo gradually becomes evident, and its response speed in later turns becomes substantially faster than that of FastV. The right panel of Fig.~\ref{fig6} further shows that VisCo surpasses FastV in terms of average response time after only three dialogue turns.

Overall, VisCo consistently reduces KV cache memory across all evaluated scenarios, and demonstrates clear efficiency advantages in long-response and single-image multi-question scenarios.

\section{Conclusion}
In this paper, we present VisCo, a training-efficient visual token compression framework that leverages the pretrained VLM itself as an intrinsic compressor. By performing compression through lightweight adaptation while keeping the backbone intact, VisCo avoids the severe degradation of training-free methods under aggressive compression and the heavy retraining cost of external-module-based approaches. Extensive experiments on 3 VLM backbones and 6 benchmarks show that VisCo consistently surpasses existing methods across diverse compression ratios, while remaining effective even in the extreme one-token setting. Overall, VisCo demonstrates that leveraging the intrinsic priors of pretrained VLMs enables robust visual token compression with only lightweight training under tight token budgets. Future work will explore adaptive token allocation for different visual inputs and extend VisCo to more challenging video compression settings.

\begin{acks}
This work is supported by the National Natural Science Foundation of China (Grant No. 62272430).
\end{acks}

\bibliographystyle{ACM-Reference-Format}
\bibliography{references}

\end{document}